\documentclass{article}

\usepackage[preprint,nonatbib]{modified_neurips_2023}

\usepackage{amsmath, amssymb, amsfonts}
\usepackage{bbm}
\usepackage{dsfont}
\usepackage{xcolor}
\usepackage[bottom]{footmisc}
\usepackage{tabularx}
\usepackage{tablefootnote}
\usepackage[backend=biber,style=nature,natbib=true]{biblatex}
\bibliography{_citations}
\usepackage[verbose=true,letterpaper,tmargin=1in,bmargin=1in,left=1.15in,right=1.15in]{geometry}
\usepackage[breaklinks]{hyperref}   %
\usepackage[noabbrev,capitalize]{cleveref}
\usepackage[bottom]{footmisc}
\usepackage{graphicx}
\usepackage{caption}
\usepackage{subcaption}
\usepackage{xspace}
\usepackage[normalem]{ulem}
\usepackage{multirow}
\usepackage{float}
\usepackage{ulem}
\usepackage{multicol}
\usepackage{enumitem}
\usepackage{cuted}
\usepackage{url}
\usepackage{setspace}
\usepackage{fancyhdr}

\crefformat{footnote}{#2\footnotemark[#1]#3}
\def\eqref#1{equation~\ref{#1}}

\def\1{\bm{1}}

\DeclareMathAlphabet{\mathsfit}{\encodingdefault}{\sfdefault}{m}{sl}
\SetMathAlphabet{\mathsfit}{bold}{\encodingdefault}{\sfdefault}{bx}{n}

\newcommand{\ifprecedingtext}[1]{\ifvmode\relax\else#1\fi}

\newcommand{\si}{semantic interpretation\xspace}
\newcommand{\Si}{Semantic interpretation\xspace}
\newcommand{\SI}{Semantic Interpretation\xspace}
\newcommand{\mi}{algorithmic interpretation\xspace}
\newcommand{\Mi}{Algorithmic interpretation\xspace}
\newcommand{\MI}{Algorithmic Interpretation\xspace}

\newenvironment{blue}{
    \color{blue}
}{
    \ignorespacesafterend
}
\newenvironment{orange}{
    \color{orange}
}{
    \ignorespacesafterend
}
\newenvironment{olive}{
    \color{olive}
}{
    \ignorespacesafterend
}

\title{
The Cognitive Revolution in Interpretability: \\
From Explaining Behavior to Interpreting Representations and Algorithms %
}

\author{%
  Adam Davies \\
  Department of Computer Science \\
  University of Illinois Urbana-Champaign \\
  \texttt{adavies4@illinois.edu} \\
  \And
  Ashkan Khakzar \\
  Department of Engineering Science \\
  University of Oxford \\
  \texttt{ashkan.khakzar@eng.ox.ac.uk}
}

\begin{document}

\maketitle

\begin{abstract}

Artificial neural networks have long been understood as ``black boxes'':
though we know their computation graphs and learned parameters,
the knowledge encoded by these weights and functions they perform
are not inherently interpretable.
As such, from the early days of deep learning, there have been efforts to explain these models' behavior and understand them internally;
and recently, \emph{mechanistic interpretability} (MI) has emerged as a distinct research area studying the features and implicit algorithms learned by foundation models such as large language models.
In this work, we aim to ground MI in the context of cognitive science, which has long struggled with analogous questions in studying and explaining the behavior of ``black box'' intelligent systems like the human brain.
We leverage several important ideas and developments in the history of cognitive science to disentangle divergent objectives in MI and indicate a clear path forward.
First, we argue that current methods are ripe to facilitate a transition in deep learning interpretation echoing the ``cognitive revolution'' in 20th-century psychology that shifted the study of human psychology from pure behaviorism toward mental representations and processing.
Second, we propose a taxonomy mirroring key parallels in computational neuroscience to describe two broad categories of MI research, \emph{\si} (what latent representations are learned and used) and \emph{\mi} (what operations are performed over representations) to elucidate their divergent goals and objects of study. %
Finally, we elaborate the parallels and distinctions between various approaches in both categories, analyze the respective strengths and weaknesses of representative works, clarify underlying assumptions, outline key challenges, and discuss the possibility of unifying these modes of interpretation under a common framework.
\end{abstract}

\section{Introduction}\label{sec:intro}

How can we understand, interpret, and explain the behavior of complex intelligent biological systems like humans, chimpanzees, dolphins, or (at least seemingly) intelligent AI systems like large language models (LLMs)?
This question lies at the heart of cognitive science, and different answers have defined entire paradigms in member disciplines such as psychology, neuroscience, and linguistics.
For instance, in the 1950s and early 1960s, the dominant paradigm in academic psychology was behaviorism \cite{mandler2002origins}, resting on the fundamental assumption that human behavior can be fully understood and explained in terms of stimulus-and-response mechanisms \cite{skinner1957verbal,chomsky1980review}, with no need to consider humans' internal mental states, representations, or processing.
However, behaviorism failed to account for many key facets of human psychology, including language acquisition \cite{chomsky1980review,chomsky2002syntactic}, concept learning and problem-solving \cite{bruner2017thinking}, and working memory \cite{miller1956seven}, leading to the so-called ``cognitive revolution'' of the 1960s that birthed the modern cognitive sciences built around studying human cognition on the basis of mental representation and information processing \cite{sperry1993revolution,miller2003revolution}.

Most contemporary deep learning research also rests upon the behaviorist assumption -- i.e., that it is only necessary to study models' behaviors (and not their internal representations, processing, etc.) in order to understand and explain their capabilities, limitations, benefits, and risks \cite{raji2021benchmark,mahowald2023dissociating}.
However, recent developments in 
mechanistic interpretability have challenged this paradigm, opening the door for a potential cognitive revolution in deep learning. 
The goal of such work is not (only) to explain specific model \emph{outputs}, but more broadly to interpret the latent representations \cite{rogers2020bertology,belinkov2022probing,templeton2024scalingmonosemanticity} or internal operations and algorithms \cite{elhage2021circuit,conmy2023circuit} that are learned in self-supervised pretraining, and to explain how these representations and mechanisms result in observed model behaviors \cite{elazar2021amnesic,davies2023calm,geiger2023causal}.
Specifically, we break this work into two broad categories:\footnote{
    In this categorization, we substitute the more common term ``interpretability'' with ``interpretation'', as each category of work is concerned with \emph{interpreting} the inner structure of an otherwise ``black box'' model, rather than the state of being inherently interpretable (as studied in the area of interpretable machine learning; see \cref{sec:olddfns}).
}
\begin{itemize}
    \item \emph{\Si}: what latent \emph{representations} are learned and used by models? (See \cref{sec:si}.)
    \item \emph{\Mi}: what \emph{operations} and \emph{algorithms} are being implicitly implemented by models (over such representations) to carry out a given behavior? (See \cref{sec:mi}.)
\end{itemize}
Our goal is to explore the relationship between semantic and \mi, elaborating the different assumptions made by each category of work and how they each approach the notion of interpretation, discussing the parallel challenges faced by each category, and grounding such work in concrete parallels from the history of cognitive science in order to better understand how associated cognitive disciplines have approached the question of understanding, interpreting, and explaining the behavior of intelligent systems.

\section{Background}\label{sec:bg}

\subsection{A Brief History of Deep Learning Interpretation}\label{sec:interpret-explain}

Since their introduction, neural networks have been considered "black boxes,"\footnote{
    Note that, by ``black box'' model, we are not simply referring to models that are only accessible via APIs such as ChatGPT; we refer to all models that are not \emph{transparent}, meaning that the operations being performed by the model to transform inputs into outputs cannot be easily interpreted by human practitioners simply by inspecting the model (see \cref{sec:olddfns}). This includes all but the very simplest neural networks.
} meaning they are not inherently interpretable.
The research on deep learning interpretation is concerned with casting light on the ``black box'' problem in some form or another.
In the era of deep learning preceding the development of foundation models (i.e., neural networks pre-trained on large-scale, self-supervised tasks such as LLMs \cite{bommasani2021foundation}), the terms \emph{interpretation} and \emph{explanation} most often referred to saliency (feature attribution) methods. 
However, following the paradigm shift toward foundation models, these terms now more often refer to mechanistic interpretation. 
In this section, we provide a broad overview of both periods and associated paradigms in deep learning interpretation.

\paragraph{Behaviorist Interpretation: The Rise of Post-hoc Explanations}\label{sec:behaviorist}
The paradigms of behaviorist and mechanistic interpretation of neural networks emerged concurrently, shortly following the modern study of deep learning.
For instance, early work in deep learning interpretation discussed explaining the network behavior by identifying the importance of input features for the output, coining the term ``saliency maps'' \cite{simonyan2014saliency}; and contemporaneous work studied representations by analyzing different neuron activations within convolutional neural networks \cite{zeiler2014visualizing}. 
However, the focus on post-hoc behavioral explanations -- i.e., explaining why a given model produced a specific outputs \cite{lipton2018mythos,doran2017xai,arrieta2020xai} -- gained more traction in the context of the then-dominant paradigm of classification and supervised learning,
particularly in the forms of feature attribution methods and counterfactual explanations. 
Specifically, feature attribution \cite{SHAPLundberg2017,IntegratedGradAxiomSundararajan2017,ManyShapleySundararajan2020,LIMERibeiro2016,fong2019understanding,SaliencySimonyan,GradCAM,khakzar2022ModelKnowsBest,LRPbach2015pixel} analyzes the behavior of the network by identifying input features that are relevant for that output;
and counterfactual explanations \cite{wachter2017counterfactual,lang2021explainingInStyle,checklist} understand the behavior of the network by answering what needs to change in the input for the network to have a particular output. 
We highlight that these methods are analogous to the behaviorist paradigm in psychology (discussed in \cref{sec:intro}), which studied cognition only in terms of observable behaviors, not internal mental representations or cognitive processing.

\paragraph{Mechanistic Interpretation: The Cognitive Revolution in Interpretability}\label{sec:cogrev}

With the rise of generative and self-supervised learning paradigms that have enabled increasingly powerful and task-general foundation models, the focus has shifted toward questions regarding what these models actually learn in pre-training and what implicit algorithms they perform internally, rather than why a model generated a particular output in a downstream task.
Such work is often referred to as \emph{mechanistic interpretability}, which is generally defined as the subfield of interpretability research concerned with ``reverse engineer[ing] neural networks, similar to how one might reverse engineer a compiled binary computer program'' \cite{olah2022mechanistic}.
This description clearly covers implicit algorithms and individual constituent operations, as studied in circuit discovery (see \cref{sec:mi}); but it is less clear precisely how mechanistic interpretability relates to the study of latent representations of human-interpretable concepts, particularly when specific concepts of interest are provided in advance of empirical analysis (as in probing; see \cref{sec:sp,sec:cp}), rather than dynamically discovered from embedding representations (as in dictionary learning; see \cref{sec:sae}).
In order to avoid conflating these related but fundamentally distinct notions of interpretability and clarify the specific object of analysis for each family of work, we define a taxonomy of interpretation methods according to the study of latent representations (\emph{\si}; see \cref{sec:si}) and operations/algorithms (\emph{\mi}; see \cref{sec:mi}), and discuss the possibility of a unified interpretation framework integrating both perspectives in \cref{sec:unified}.

\paragraph{Complementary Paradigms} It is important to recognize that mechanistic interpretation and behavioral explanation are complementary rather than competing. 
For instance, understanding why a given deep neural network produced a certain output remains crucial, especially in safety-critical domains such as healthcare \cite{reddy2022medicine,petch2022opening,thirunavukarasu2023medicine}; and monitoring the latent knowledge and capabilities of frontier models has key implications for AI safety \cite{hendrycks2021unsolved,zou2023representation}. Such directions are deeply intertwined and should continue to inform each other, as they have throughout the history of interpretability. For instance, internal representations have been leveraged for feature attribution explanations in Class Activation Mapping (CAM) \cite{zhou2016learningCAM}, where neuron activations are used to explain which input features are relevant to the output; and subnetwork analysis (which is closely related to circuits; see \cref{sec:mi}) has worked to explain behaviors in terms of input features by finding and ablating sparse internal pathways through the network \cite{LRPbach2015pixel,khakzar2021CriticalPathway}. 
Most recently, causal probing \cite{elazar2021amnesic,davies2023calm} and dictionary learning \cite{bricken2023monosemanticity,templeton2024scalingmonosemanticity} have been leveraged for explaining the behavior of LLMs in terms of interpretable latent features (see \cref{sec:cp,sec:sae}, respectively), and circuit discovery has been applied to uncover the interpretable subnetworks of LLMs that correspond to certain behaviors \cite{wang2023ioi,conmy2023circuit} (see \cref{sec:mi}). 
This interplay mirrors parallel approaches to understanding human intelligence as observed between behavioral and cognitive modes of analysis: just as cognitive neuropsychology can be invaluable in explaining real-world behaviors and pathologies, mechanistic interpretations can inform our understanding of model outputs, and vice versa. Both perspectives are essential for a holistic understanding of neural networks.

\subsection{Levels of Analysis}\label{sec:marr}
Marr's levels of analysis \cite{marr1982levels} have long been a workhorse foundational framework for scientific inquiry in neuroscience and other disciplines of cognitive science \cite{niv2016levels}. They are as follows:
\begin{itemize}
    \item The level of \textbf{computational theory} provides a mathematical description of the \textbf{goal} of an information-processing system in terms of the desired transformation from inputs to outputs.  %
    \item The level of \textbf{representations and algorithms} is concerned with how the system represents inputs and outputs, and the algorithm it employs to carry out the transformation described by the computational theory.  %
    \item The level of \textbf{implementation} is concerned with how representations and algorithms are realized in a physical or software medium.  %
\end{itemize}

A few prior works have analyzed the role of Marr's levels of analysis in the context of machine learning, approaching these levels from the perspective of the learning process \cite{hamrick2020levels,sawant2020levels}.
For example, \citet{hamrick2020levels} consider the scenario of Deep Q-Networks \cite{mnih2015deepq}, formulating the computational level as the task of mapping input observations to output actions that maximize a given reward function, the algorithmic level as the learning algorithm used to train the network,\footnote{
    For each example provided by \citet{hamrick2020levels}, at the level of representations and algorithms, only algorithms are considered. This is natural outside the context of interpretability, as it is not possible to directly interpret representations from dense neural embeddings, meaning that suitable \si methods are a prerequisite for studying representations in addition to algorithms at this level.
} and the implementation level as the set of choices for implementing this network in software (e.g., the network architecture, hyperparameters, optimizer, etc.).
Useful as such examples may be for interrogating various design decisions in training such a network, they are orthogonal to the question of what is actually being represented by the network or what implicit algorithms are being learned to support its task performance.
Indeed, only trivial applications of Marr's levels to a pre-trained network itself (rather than simply the learning process that produced it) are possible without some notion of interpretation: otherwise, one can only indicate embedding vectors as representations and the forward pass of the network as the algorithm, precluding any analysis of latent properties that are being represented by embeddings and the implicit algorithm that is being carried out in the forward pass.

However, as we will show in the following sections, recent developments in semantic and \mi have enabled the study of these models (and not only the learning process that produced them) at the level of representations and algorithms, where studying representations is a matter of \emph{\si}, and studying algorithms is a matter of \emph{\mi} (discussed in \cref{sec:si,sec:mi}, respectively).

\section{\SI}\label{sec:si}
We may define \emph{\si} %
as a matter of answering the following question: what latent properties are learned and represented by neural networks, and how do they contribute to observed model behaviors?
For example, do vision models learn representations of semantically-related object categories or spatial relations; or do LLMs learn representations of 
syntactic dependencies or lexical relations?

\subsection{Optimization and Search}
\paragraph{What inputs maximize the activation of a given neuron?}
Given that individual neurons are the atomic level of representation for all neural networks, a natural question is, to what extent do single neurons code for distinct properties of interest?
An analogous hypothesis in neuroscience is the notion of ``gnostic cells'', cells (neurons) which fire only in the context of very specific concepts, such as one's grandmother (hence the name) \cite{gross2002gnostic,quiroga2013gnostic}. Do we find such neurons in contemporary deep learning systems?
The most prominent approach to neuron-level \si involves selecting an individual neuron and either searching through a pre-defined query set or generating an input (such as an image) that maximally activates the target neuron, and inspecting these optimized inputs for common features.
Early work in this area focused on image recognition networks \cite{erhan2009visualizing,zeiler2014visualizing,simonyan2013deep,zhou2014object},
where some approaches operate by searching for patterns that maximize the output of neurons via image optimization
\cite{simonyan2014saliency,zeiler2014visualizing,nguyen2016synthesizing,olah2017feature},
and others searched through a pre-defined query set to find inputs that maximized neuron activations \cite{erhan2009visualizing,fong2018net2vec,bau2017network,bau2018visualizing}.
For instance, Net2Vec \cite{fong2018net2vec} and Network Dissection \cite{bau2017network,bau2018visualizing} systematically analyze the relationship between concepts and neurons by analyzing the activations of neurons in response to images in the dataset clustered by human-annotated visual concepts.
Later work has investigated how the same paradigm can be applied to interpret individual neurons in language models \cite{karpathy2015visualizing,dalvi2019oneSand,geva2021keyvalue,dai2022knowledge} and multimodal vision-language models \cite{hernandez2021natural,schwettmann2023multimodal}.

\subsubsection{Assumptions and Challenges}
\paragraph{Levels of Representation}
Most ``optimization and search''-based methods operate at the level of individual neurons, which comes with serious limitations: neural networks encode \emph{distributed} representations where features are encoded by multiple neurons; and even small-scale, simplified ``toy'' models have also been found to exhibit \emph{polysemanticity}, where multiple concepts are represented by a single neuron \cite{elhage2022superposition}.
As such, the underlying assumption made in interpreting the semantics of only single, isolated neurons -- i.e., that there is a one-to-one mapping from neuron activations to values taken by latent properties (analogous to the ``gnostic cell'' hypothesis) -- is in no way guaranteed to hold.
Some works have relaxed this assumption by considering combinations of neurons instead of only individual neurons \cite{geva2021keyvalue,dai2022knowledge}, but the broader question remains: is a subset of neurons in a given layer the appropriate level at which to analyze neural representations, or should the activations of all neurons in a given layer (i.e., its embedding space) be taken into account?
Another question raised by such work is whether and how different architectures should be taken into account when examining neurons activations -- e.g., for Transformer-based models, do self-attention layers merit special consideration \cite{voita2019attention,vig2019attention,clark2019attention}, or should neurons in feed-forward layers also be examined \cite{geva2021keyvalue,dai2022knowledge,meng2022rome,meng2023editing}?

\paragraph{Needle in a Haystack}
Finally, perhaps the greatest concern with analyses centered around the activations of individual neurons (or small sets of such) is that billion-parameter scale models have millions of functional neurons, and this number will continue to scale exponentially alongside the number of parameters in state-of-the-art models. 
Naturally, it would be computationally intractable to perform detailed analysis with respect to each individual neuron; so how can one determine which neurons are meaningful and merit individual analysis, and which neurons can be ignored?
While a number of heuristic approaches have been proposed for targeting neurons of interest (e.g., see \citealt{bau2018visualizing,meng2022rome,schwettmann2023multimodal}, \emph{inter alia}), there is currently no generally accepted methodology for determining which neurons are most important for any given analysis, meaning that there is no way to guarantee (or even provide bounds on the probability) that one has correctly targeted the neurons whose activations are most important in studying any given research question.

\subsection{Structural Probing}\label{sec:sp}

\paragraph{Framework}
One of the most popular and well-studied approaches to %
has been \emph{structural probing} \cite{belinkov2022probing,rogers2020bertology,belinkov2020probetypes}. The goal of structural probing is to train auxiliary classifiers (probes) to predict discrete latent properties of inputs from model embeddings. 
For example, one may train a probe to predict parts-of-speech from LLM token embeddings, so when given each token in the sequence (\texttt{The}, \texttt{cat}, \texttt{meows}, \texttt{for}, \texttt{dinner}), the probe predicts the corresponding parts of speech (\texttt{determiner}, \texttt{noun}, \texttt{verb}, \texttt{preposition}, \texttt{noun}), respectively.
A strong form of the underlying assumption here is that a model is ``representing'' a property if and only if this property can be consistently predicted from embeddings (e.g., if probes can achieve high validation accuracy w.r.t. to the property in question).
For example, an early and influential argument in structural probing, the \emph{pipeline hypothesis}, uses probe accuracies over linguistic tasks across BERT \cite{bert} layers to argue that BERT processes linguistic properties in the same order as the ``classical NLP pipeline'', with surface-level features recognized first, followed by syntactic features, and semantic features recognized last \cite{tenney2019pipeline,jawahar2019pipeline,rogers2020bertology,niu2022pipeline}. %

\subsubsection{Assumptions and Challenges}\label{sec:siassume}

\paragraph{Levels of Representation}\label{sec:siarch}
In order to carry out structural probing research, one must first define the class of representations to be probed -- or equivalently, the architecture of the probe being trained (e.g., linear probes can only detect properties that are linearly-encoded).
The question of which probing architecture should be utilized is a contentious one \cite{belinkov2022probing,hewitt2019controlprobe,pimentel2022architectural};
and below, we outline several choices of probing architecture as utilized in the literature, supporting arguments in favor of each architecture, and corresponding limitations.

\textit{\underline{Linear Probes}}\hspace{0.5em}
Perhaps the most well-studied probing architecture is the \emph{linear probe} \cite{alain2016linear,kim2018cav,liu2019linguistic,ravfogel2020inlp,elazar2021amnesic,schwettmann2021toward,park2023linear,marks2023geometry,tigges2023linear,nanda2023linear}.
Use of such probes assumes the \emph{linear subspace hypothesis}: that property representations are encoded by linear embedding subspaces \cite{vargas2020linear,bolukbasi2016debiasing}. 
For instance, Concept Activation Vectors (CAV) \cite{kim2018cav} are vectors in the representation space perpendicular to a linear classification boundary that classifies the representations of a concept versus other input representations.
One argument in favor of linear probing is the intuition that neural classifiers must make class-discriminative information linearly separable in their final embedding layer, so probes (particularly over final-layer embeddings) should also be linear \cite{alain2016linear}.
Another motivation is that, given enough training data, sufficiently expressive probes can memorize arbitrary probe tasks irrespective of the model being probed \cite{hewitt2019controlprobe}, so the accuracy of a simpler (i.e., less expressive) probe may better reflect the actual content of embeddings rather than the expressiveness of the probe.

\textit{\underline{Nonlinear Probes}}\hspace{0.5em}
An alternative approach to structural probing involves training arbitrarily expressive (nonlinear) probes in order to learn any representation that may be encoded by the model \cite{pimentel2020information,white2021nonlinear,pimentel2022architectural,davies2023calm}.
For instance, \cite{crabbe2022concept} extends CAV using kernels to classify concept regions instead of only concept vectors.
As neural networks are, by design, highly nonlinear mathematical objects, it is natural to expect that they may encode some properties nonlinearly \cite{white2021nonlinear}. This is particularly true of earlier or intermediate layers, where -- unlike the final layer -- class-discriminative information does not need to be made linearly separable in order to facilitate classification.
An additional argument in favor of nonlinear probes is that probes should mirror the architecture of the model being probed, as this more directly reflects the information that is usable by the model \cite{pimentel2022architectural}.
However, as noted above, there have been concerns that highly expressive probes may memorize the mapping from embeddings to properties irrespective of the model being probed \cite{hewitt2019controlprobe}; so for more complex probes, there is an increased risk that high probing accuracy is more a reflection of the probe itself than it is of the model being probed.
For instance, in an extreme case, consider 
generative vision-language models that use embeddings from a frozen image encoder and fine-tune an LLM to generate corresponding image captions \cite{zhang2021vinvl,zhai2022lit}. In this case, the LLM
could be understood as a highly complex and expressive probe in the sense that the LLM is an auxiliary network (cf. probe) stacked on top of a frozen vision encoder (cf. model being probed) and trained to 
generate text captions (cf. probe task).
In this case, it is clear that the generative ``probe task'' is, in fact, being learned by the probe (LLM) and not the original model (vision encoder), given that the vision encoder is never trained to generate text.

\paragraph{Probing for what?}
Another important assumption is the choice of properties for which to probe.
It is impossible to know \emph{a priori} which properties a model is representing or leveraging in any given context \cite{rogers2020bertology,yun2021dictionary,belinkov2022probing}; so for any given probing experiment, it is always possible that one has simply failed to capture whatever properties are most important to the model, leading to potentially misleading results.
This is a serious concern for \si, given that we cannot reasonably presume to know ahead of time what complete set of properties may be represented and leveraged by models or whether they happen correspond to key properties in human cognition, and there is a long-documented trend toward anthropomorphizing intelligent-seeming models (especially in the context of linguistic systems such as LLMs) \cite{weizenbaum1966eliza,turkle2007authenticity,switzky2020eliza}.
For instance, while substantial early work in structural probing studied ``classic NLP pipeline'' properties \cite{tenney2019pipeline,jawahar2019pipeline,rogers2020bertology,niu2022pipeline} (as discussed above), there is no particular reason to believe that such properties are the most important features for interpreting the internal representation or explaining the behavior of any given LLM.
Additionally, there are many ways to interpret structural probing results \cite{belinkov2022probing}: naively, one might simply consult probing accuracies and compare them between properties or across layers; but various works have argued for the necessity of comparing probe predictions against randomized baselines \cite{tenney2019baseline}, control tasks \cite{hewitt2019controlprobe}, or to compute information gain using control functions \cite{pimentel2020information}.

\paragraph{Correlation does not imply causation.}
Perhaps the single most important concern with structural probing is that, under this paradigm, it is only possible to measure \emph{what properties can be predicted} from a representation, not \emph{whether (or how) they are actually used} by the model \cite{belinkov2022probing,elazar2021amnesic,hewitt2019controlprobe,davies2023calm}.
We discuss a few proposed solutions to this problem in the following sections.

\subsection{Causal Probing}\label{sec:cp}

\paragraph{Framework}
One proposed solution to structural probing's inability to distinguish correlation from causation problem is \emph{causal probing}, where interventions are performed over representations detected by (structural) probes, and the resulting impact on model behaviors is measured in order to study how these properties are used by the model \cite{elazar2021amnesic,davies2023calm,lasri2022probing}.
For instance, \emph{amnesic probing} \cite{elazar2021amnesic} uses the iterative nullspace projection (INLP) algorithm \cite{ravfogel2020inlp} to remove all information that is linearly-predictive of a given target property from LLM embeddings, then performs this intervention in the model's forward pass to measure the impact of the removal operation on language modeling performance across a large text corpus to broadly estimate and compare the model's use of various target properties. %

\subsubsection{Assumptions and Challenges}
\paragraph{Inherited from Structural Probing}
As in structural probing, causal probing requires one to pre-define the level of representation (neuron-level, linear, or nonlinear) and set of properties for which to probe before any probing experiment can begin.
Most intervention methodologies are built to operate at only a single level of representation -- typically linear \cite{ravfogel2020inlp,elazar2021amnesic,ravfogel2021counterfactual,ravfogel2022linearadversarial}, with some work exploring kernelized linear representations \cite{ravfogel2022kernel,shao2023spectral} -- meaning that methods and results from one level cannot be directly adapted or compared to those from other levels.
Alternative approaches have been defined using adversarial attacks against arbitrary probing architectures, allowing interventions to target whatever level of representation assumed by the probe \cite{tucker2021counterfactual,davies2023calm}; but these approaches come with fewer theoretical guarantees on potential collateral damage to non-targeted properties, as discussed below.

\paragraph{Completeness vs. Selectivity}\label{sec:minmax}
An important observation made by \citet{elazar2021amnesic} is that properties removed from earlier layers can be \emph{recoverable} by later ones -- i.e., 
when INLP is used to remove information about some target property from the embeddings of a given upstream layer, it is often still possible to train (linear) probes to predict the property from embeddings of a later downstream layer, meaning that these linear interventions are \emph{incomplete}.
For such properties, this finding may be taken as evidence against the linear subspace hypothesis discussed above: INLP removes \emph{all} information that is linearly predictive of the target property, so if BERT only encoded these properties linearly, it would not be possible to recover them following an INLP intervention.
Later works in causal probing have investigated nonlinear interventions \cite{ravfogel2022kernel,shao2023spectral,davies2023calm}, but as in structural probing, there is a tradeoff associated with expressivity: theoretically, the more powerful (and potentially more \emph{complete}) an intervention is,
the more ``collateral damage'' it may also cause to representations more generally \cite{zhao2022tradeoff}, in which case the intervention is less \emph{selective} in restricting damage to only the target property.\footnote{
    Here, we use \emph{selectivity} in the sense described by \citet{elazar2021amnesic}, and not other probing work such as \citet{hewitt2019controlprobe}, where it instead refers to the gap in performance between probes trained to predict real properties versus randomized ``nonsense'' properties.
}
Thus, for the most general intervention methodologies (such as those proposed by \citet{tucker2021counterfactual,davies2023calm}, which can be used to manipulate representations detected by any differentiable probe), it is difficult to determine whether any observed changes to model behavior in the presence of interventions is attributable to the model's representation of the target property or simply to collateral damage (i.e., low selectivity).

\paragraph{Rashomon Effect}\label{sec:sirashomon}
Finally, as in most studies of causality, causal probing introduces the Rashomon effect \cite{breiman2001rashomon,hancox2020rashomon}: when there are multiple explanations with equal causal efficacy 
(e.g., if removing information about property A and property B from the model leads to the same impact in its behavior), there is no way to determine which explanation is ``correct'' (e.g., we cannot say whether the model's representation of A or B is responsible its behavior) \cite{mueller2024missedcauses}.

\subsection{Dictionary Learning}\label{sec:sae}

As noted in \cref{sec:sp,sec:cp}, one of the key limitations of both structural and causal probing is that they are fully supervised, meaning that one must pre-define a set of latent properties for which to probe.
This means that, even for a perfect (causal) probing methodology, it is always possible that the most important properties leveraged by the model in the course of performing a particular task could be completely missed if one simply does not probe for these particular properties \cite{rogers2020bertology,yun2021dictionary,belinkov2022probing}.
This is a serious concern for \si, given that we cannot reasonably presume to know ahead of time what complete set of properties may be represented and leveraged by a model in any given context.

\paragraph{Dictionary Learning}
An alternative paradigm in \si, \emph{dictionary learning}, removes this presumption by inverting the traditional probing process: 
instead of directly training a \emph{supervised} probe to predict some latent property of interest from a model's intermediate embedding representations,
the goal of dictionary learning is to train an \emph{unsupervised} probe to
decompose embeddings into a sparse combination of features and use them to reconstruct the original embeddings, yielding a \emph{dictionary} of features that are useful for sparsely representing embeddings \cite{lewicki2000overcomplete,lee2006sparsecoding,faruqui2015sparse}.

\paragraph{Sparse Auto-Encoders}
Sparse Auto-Encoders (SAEs) \cite{subramanian2018spine,yun2021dictionary} have recently emerged as a powerful and scalable approach to unsupervised probing via dictionary learning \cite{cunningham2023sae,bricken2023monosemanticity,templeton2024scalingmonosemanticity}, performing a nonlinear transformation of input embeddings onto an overcomplete linear basis, allowing them to learn (potentially exponentially many) more features than the dimensionality of the embeddings they are trained on.
Where early dictionary learning methods in signal processing were based on sparse coding methods involving Bayesian modeling \cite{lewicki2000overcomplete} or convex optimization \cite{lee2006sparsecoding}, SAEs carry out dictionary learning using neural networks, improving scalability while maintaining the same goal of learning sparse features for decomposition and reconstruction.

\subsubsection{Assumptions and Challenges}

\paragraph{Levels of Representation: Superposition}
As with structural and causal probing above, it is necessary to define the level of representation (e.g., neuron-level, linear, or nonlinear) as the target of analysis.
Formally, SAEs (as formulated above) are an unsupervised \emph{nonlinear} probe that project embeddings onto an overcomplete \emph{linear} basis, and thus fall somewhere in-between the linear and nonlinear levels of representation.
Instead, SAEs follow the \emph{superposition hypothesis} \cite{elhage2022superposition}, which argues that models learn to represent more features than they have neurons in a given layer by encoding them via \emph{almost-orthogonal} directions in the embedding space, expanding the number of features that can be represented in the embedding space at the cost of some noise due to interference between non-orthogonal features.\footnote{
    Specifically, with $d$ number of neurons, embeddings can encode $\mathcal{O}(\exp(d))$ features with less than $\epsilon$ cosine similarity \cite{dasgupta2003elementary}.
}
According to this hypothesis, models leverage superposition because the benefits of representing (potentially many) more features with fewer neurons outweighs the cost of filtering out this noise (via nonlinear activation functions) so long as features are sufficiently sparse (leading to less interference on average) \cite{elhage2022superposition,scherlis2022polysemanticity}.

\paragraph{Unsupervised Feature Interpretation}
While SAEs (and dictionary learning more broadly) remove the need for labeled training data for supervised probes, they effectively transfer the burden of annotation from probe training data to unsupervised feature interpretation, as each feature vector in the dictionary cannot be directly interpreted any more easily than dense embeddings themselves.
Rather, each feature must be retroactively interpreted, usually by asking an annotator (either a human or an LLM \cite{bills2023explainneurons}) to inspect the input samples that maximally activate the feature and annotating it according to whatever feature these samples intuitively appear to have in common \cite{cunningham2023sae,bricken2023monosemanticity,templeton2024scalingmonosemanticity}.
For instance, in the largest SAE study to date (including up to 34 million SAE features learned from embeddings of a frontier LLM), \citet{templeton2024scalingmonosemanticity} find that one feature learned by an unsupervised probe over a large multilingual, multimodal vision-language model corresponds to the \emph{Golden Gate Bridge}, and that this feature is strongly activated in contexts discussing the bridge in many different languages or for images depicting the bridge, while being weakly activated in contexts discussing (or images depicting) other tourist landmarks in San Francisco or other famous bridges.
While such features are indeed interesting, it is not clear what proportion of features have such clear interpretations -- 
how many of these millions of features are as easily interpreted as ``Golden Gate Bridge''?
At such a large scale, it is not feasible for human annotators to manually interpret features, requiring automated interpretation -- e.g., prompting another LLM to explain the relationship between multiple passages that highly activate the same feature, a scalable approach that has been shown to be reasonably well-aligned with human judgments \cite{bills2023explainneurons}.
However, it is important to note that, in contrast to supervised probing methods (where target properties are labeled in advance), unsupervised feature interpretations (whether human- or LLM-annotated) are not \emph{transferable} in that this process must be performed \emph{each time} one analyzes a new model, layer, or trains a new SAE -- that is, where probing datasets only require that each input (or parts of inputs, such as individual tokens) be labeled once, dictionary learning requires that one (re-)interpret learned features every time a new dictionary is learned, significantly increasing the burden of annotation and preventing direct comparisons between different models, layers, or dictionary learning methods.

\paragraph{Evaluating Interpretations}
More broadly, even for the most intuitive features, it is not clear precisely how one should evaluate whether any given feature interpretation is correct. At the scale of frontier models and large SAEs, beyond requiring automating feature interpretation, it is also necessary to automate \emph{evaluation} of these interpretations. Thus, even for scalable and high-quality automated techniques, stacking multiple layers of LLM automation for interpretation and evaluation (which may even carried out by the same type of model that is the object of analysis) could lead to compounding errors.
Current research has addressed the problems inherent in this paradigm by leveraging \emph{causal interventions} over discovered features \cite{bricken2023monosemanticity,templeton2024scalingmonosemanticity} (directly analogous to those performed in causal probing over supervised features) in order to observe whether model behavior changes consistently with the hypothesized interpretation of each feature -- e.g., when the Golden Gate Bridge feature discussed above is significantly strengthened, the associated LLM responds to many queries with references to the famous bridge where it normally would not.
However, as in causal probing, such interventions can only tell us whether a given feature contributes to a particular output, not whether there are other features that contribute just as strongly to the prediction \cite{mueller2024missedcauses}, nor the extent to which interventions are \emph{complete} (meaning they fully control all aspects of the target feature) or \emph{selective} (meaning they do not also damage non-targeted features).

\section{\MI}\label{sec:mi}

We may define \emph{\mi} %
as a matter of answering the following questions: what operations is a given neural network implicitly performing (over representations); what role do these operations play in observed behaviors; and, when these operations are taken in aggregate, what algorithm is being implicitly implemented by the model?
Such operations and algorithms are typically understood in terms of \emph{circuits}: ``sub-graphs of the network'' (weights) that implement a given operation or algorithm, which may be themselves composed to form larger circuits \cite{olah2020circuit}.
In principle, circuits can refer to any sub-graph of a neural network; 
but \mi studying Transformer-based models (the architecture of most modern LLMs and many other foundation models) typically studies circuits at the level of individual attention heads, or compositions of such \cite{elhage2021circuit,wang2023ioi}.

\subsection{Circuit Discovery}\label{sec:circuit}

\paragraph{Circuit Discovery}\label{sec:discovery}
The task of finding circuits that faithfully describe a given category of model behaviors is often referred to as \emph{circuit discovery} \cite{wang2023ioi,conmy2023circuit}.
The simplest circuits to understand are \emph{end-to-end circuits}, which describe a full path (composed of sub-circuits) from input to output, implementing a complete algorithm.
For instance, one of the first circuits identified in a Transformer language model is the \emph{induction circuit} \cite{elhage2021circuit,olsson2022induction}: an end-to-end circuit which, given an input of the form ``[A] [B] ... [A]'' (where [A] and [B] are arbitrary token sequences), determines whether or not to predict that [B] once again follows the second [A] (as it did earlier in the sequence). For instance, given the input ``Vernon Dursley and Petunia Durs'' (tokenized into [Vern, \#on, Durs, \#ley, and, Petunia, Durs], where \# denotes a token that has been created by splitting an existing word into multiple word pieces), an induction circuit would be tasked with predicting whether or not to follow the second ``Durs'' token with ``\#ley''. (See \citet{conmy2023circuit} for additional examples of end-to-end circuits that have been discovered in LLMs.)

\paragraph{Causal Interventions}
As in causal probing and dictionary learning above, a popular strategy in circuit discovery is to explain a circuit's contribution to a model's behavior by intervening in its forward pass and study the effect on the model's predictions.
There are two popular strategies for doing so, \emph{knockouts} and \emph{patching}, which we discuss in turn below.

\textit{\underline{Knockout}}\hspace{0.5em}
Knockouts ``turn off'' a (sub-)circuit by nullifying it in order to determine its contribution to LLM behaviors.
Given a circuit, a few approaches to performing knockout have been proposed: \emph{zero ablation} sets the output of the analyzed circuit to zero \cite{olsson2022induction}; whereas \emph{mean ablation} instead sets its output to the mean value of a given reference distribution \cite{wang2023ioi}. The former approach is simpler, as it does not require one to define a reference distribution; but as subsequent circuits may ``rely on activation value[s] as an implicit bias term'' \cite{wang2023ioi}, it is theoretically more sound (and empirically less noisy) to perform knockout by setting to the mean value of a reference distribution where possible.

\textit{\underline{Patching}}\hspace{0.5em}
Patching replaces circuit outputs in the course of computing a \emph{target sequence} with activations from computing a \emph{source sequence}, allowing one to measure whether LLM outputs 
for the patched target sequence change consistently with the hypothesized function of the circuit in the context of the source sequence \cite{meng2022rome,wang2023ioi,conmy2023circuit,meng2023editing}. %
For instance, in the earlier example of an induction circuit given a source sequence of ``Vernon Dursley and Petunia Durs'', where the next token predicted by the LLM will be ``\#ley'', one may patch the circuit outputs from this source sequence into the LLM in its forward pass while processing the target sequence ``Thus, the argument must be val'' in order to swap its prediction for the next token from ``\#id'' to ``\#ley''.

\subsubsection{Assumptions and Challenges}\label{sec:miassume}

\paragraph{Circuit Architecture}
As in \si, circuit discovery requires one to target a specific level at which to interpret models
-- for instance, in probing, this is determined by the expressivity of the selected probe architecture; but in circuit discovery, one must decide on the \emph{atomic} (smallest-scale) sub-graphs considered as possible features or (sub-)circuits. For example, \citet{olah2020circuit} starts at the level of individual neuron activations; but given the intractability of considering all neurons in larger models, more recent work targeting LLMs has begun at the level of attention heads \cite{conmy2023circuit,wang2023ioi}.
While such tradeoffs are necessary in order to study models at scale, it is important to remember that any operations or algorithms implemented by sub-graphs below or outside the scope of the atomic level -- e.g., any analysis of Transformer models built exclusively on attention heads will miss sub-circuits implemented by MLPs \cite{geva2021keyvalue,meng2022rome}.

\paragraph{One-to-One Mapping}
Another important assumption in circuit discovery is that atomic sub-graphs are understood as performing one and only one operation.
Given that even the simplest of models are known to often represent multiple properties using the same neuron \cite{elhage2022superposition}, the assumption that single neurons of LLM architectures can be neatly discretized into individual, atomic operations is suspect; and larger sub-graphs can be difficult to precisely localize and may contain redundant elements \cite{shi2024hypothesis}.
This assumption can be somewhat attenuated with the interventions discussed above, as they can be used to test the extent to which knocking out or patching a given sub-graph leads to the behavior predicted by the hypothesized circuit, evaluating whether or not the sub-graph actually performs the indicated operation.
However, this process only solves part of the problem: while it can test whether or not the sub-graph is indeed involved in implementing the observed behavior, it cannot determine whether there are other sub-graphs that may also have a similar effect \cite{zhong2023pizza,mueller2024missedcauses}; and in some cases, knocking out randomized circuits can have a similar effect as knocking out circuits that are precisely calibrated to the target behavior \cite{shi2024hypothesis}.

\paragraph{Pre-Specification}
Another important limitation of current circuit discovery work is that, to our knowledge, all circuits that have so far been identified in real-world LLMs\footnote{
    I.e., LLMs that are not small-scale ``toy models'' trained specifically for the purpose of \mi studies, as in, e.g., \citet{elhage2021circuit,olsson2022induction,elhage2022superposition}.
} have required pre-specifying a full algorithmic description before they can be found.
While this significantly reduces computational complexity and thus allows discovery to scale to LLMs \cite{conmy2023circuit},
it also means that such discovery can generally only move ``top-down'': in this case, one cannot discover a circuit that implements an algorithm which has not been explicitly specified ahead of time, and will be less likely to discover intermediate sub-graphs that do not already fit into pre-specified algorithms (a more ``bottom-up'' approach that has been employed in smaller-scale ``toy model'' investigations \cite{elhage2021circuit,nanda2023clock,zhong2023pizza}).

\paragraph{Rashomon Effect}
Finally, perhaps the greatest challenge in circuit discovery is that it is possible to yield multiple distinct circuit descriptions for the same LLM behavior depending on how circuit analysis is carried out \cite{zhong2023pizza,mueller2024missedcauses}.
(another instance of the Rashomon effect discussed above). For instance, \citet{zhong2023pizza} explore an ``algorithmic phase transition'' experiment where a form of patching is used to interpolate between source and target circuits, allowing them to characterize how correctly each circuit describes the internal representation as the balance is shifted between the source and target, finding that,
in some cases, it appears that a single model may actually be performing the algorithms associated with each circuit simultaneously.

\section{Towards Unifying Semantic and Algorithmic Interpretation}\label{sec:unified}
Finally, a few theoretical frameworks have been proposed to simultaneously account for the role of representations and algorithms in model behaviors \cite{geiger2023causal,davies2023calm} -- i.e., to unify \si and \mi under a common framework.
\citet{geiger2023causal} proposes a causal abstraction formalism to interpret neural network behaviors in terms of an underlying graphical causal model, where representations are nodes in the causal model and operations over these representations are edges, and measures the faithfulness of the causal model as an explanation of the neural network by computing the instance-level alignment between the network and the causal model under interchange interventions \cite{iia}; and \citet{davies2023calm} reformulates this framework at the level of concepts instead of individual instances, extending it to concept-level interventions such as those in causal probing or dictionary learning.
Several algorithms have been proposed to empirically deploy causal abstraction frameworks to explain the behavior of large-scale neural networks such as LLMs by learning linear rotational interventions over embedding spaces to perform interchange interventions \cite{geiger2024das,wu2024boundlessdas} or substituting interchange interventions for gradient-based attacks against nonlinear causal probes \cite{davies2023calm}.
However, to date, each of these works exhibit serious empirical limitations: they either (1) consider only small-scale, simplified ``toy'' models and tasks \cite{geiger2023causal,geiger2024das};
or (2) intervene only in a single neural network layer \cite{davies2023calm,wu2024boundlessdas}; meaning that their ability to describe how real-world models carry out operations that transform representations from layer to layer -- i.e., to accomplish \mi\xspace-- has not yet been empirically demonstrated.

\section{Conclusion}
In this work, we discussed the parallels between such work and research trends in the history of cognitive science, including Marr's levels of analysis and the cognitive revolution that 
birthed the modern cognitive sciences. 
We explored the relationship between two broad categories of deep learning interpretability research, \emph{\si} (what latent properties are represented) and \emph{\mi} (what operations are performed over representations), highlighting the importance of causal analysis across both categories in delivering faithful explanations of model behavior.
For each category, we surveyed salient works, analyzed their comparative strengths and weaknesses, clarified underlying assumptions, and outlined key challenges; and we discussed the relationships between these categories of work, focusing on the parallel challenges they each face and how they can compliment each other, concluding by outlining recent efforts to provide a unified framework that can account for both semantic and \mi.
Our goal is to facilitate more open, productive conversation of deep learning interpretation by providing a common lexicon of goals, assumptions, and challenges associated with different modes of interpretation, stimulating further research toward more rigorous neural network interpretation, and informing current discourse by consulting lessons from the cognitive sciences.

\newpage
\printbibliography

\appendix

\section{Supplementary Background}

\subsection{Interpretable Machine Learning}\label{sec:olddfns}
Many definitions of ``interpretability'' have been proposed to describe methods that fall outside our analysis in this work \cite{doshi2017rigorous,lipton2018mythos,arrieta2020xai,broniatowski2021nist} concerning either (1) interpretability as an inherent feature of human-understandable methods, or (2) input-output explanations (e.g., to what input features is a given output attributable) of \emph{supervised} deep learning models. 
In contrast, in this work we are concerned with the study of internal mechanisms and latent representations learned by self-supervised (foundation) models that has come to characterize much of the interpretability landscape, and is now most often referred to under the broad umbrella of \emph{mechanistic interpretability} (see \cref{sec:cogrev}).

For example, the most widely-cited definition of interpretability is provided by \citet{lipton2018mythos}, who provides two general categories of interpretability. The first is \emph{transparency}, an inherent quality of models that can be fully decomposed into clear, comprehensible operations (e.g., decision trees or rule-based expert systems). The second is \emph{post-hoc explanations}, a family of techniques to explain specific outputs of an otherwise opaque (``black box'') model (e.g., saliency maps \cite{simonyan2014saliency}).
Naturally, any interpretability work studying neural networks must fall into the category of post-hoc explanations, as neural networks are inherently opaque.
However, the notion of post-hoc explanation is, on its own, insufficient to describe the wealth of interpretability research that has developed around foundation models: the categories of post-hoc explanation techniques outlined by \citet{lipton2018mythos} predate the era of self-supervised foundation models that have come to dominate many areas of study in AI and ML, and do not apply to contemporary internal analysis techniques such as 
probing (see \cref{sec:sp})
or circuit discovery (see \cref{sec:mi}), where the goal is not necessarily to explain specific model outputs, but rather to interpret the internal representations, operations, and algorithms that foundation models learn in self-supervised pretraining.
While several works have aimed to update or refine \citet{lipton2018mythos}'s interpretability taxonomy and definitions in various ways (e.g., see \citet{arrieta2020xai,doran2017xai,adadi2018blackbox}), 
none of these have addressed the disconnect between \emph{post-hoc explanations} as a matter of explaining specific outputs, and the abundance of recent work oriented around the broader, more internal notion of interpretability that describes the study of latent representations and operations learned by foundation models.

\end{document}